\documentclass[10pt, a4paper]{article}

\usepackage[final]{lrec2026} %

\usepackage{latexsym}
\usepackage{microtype}
\usepackage{booktabs}

\title{MorfFlex: Handling Rich Morphology}

\name{Jaroslava Hlaváčová, Marie Mikulová, Barbora Štěpánková, \\
\large\textbf{ Milan Straka,  Jan Hajič}} 

\address{Charles University, Faculty of Mathematics and Physics, Institute of Formal and Applied Linguistics \\
         Malostranské náměstí 25, 118 00 Prague 1, Czech Republic \\
        \{hlavacova,mikulova,stepankova,straka,hajic\}@ufal.mff.cuni.cz\\}

\abstract{
We present MorfFlex, a morphological dictionary architecture suitable for languages with extensive regularity in both inflection and derivation. As the primary example of MorfFlex in use we introduce MorfFlex CZ, a morphological dictionary of Czech.
It is distributed as a simple, unstructured list of <wordform, lemma, tag> triplets, however, its manually maintained, unpublished source files and conversion scripts encode a sophisticated system of inflectional and derivational patterns. These patterns dramatically reduce the otherwise enormous size of the dictionary, which currently contains over 100 million wordforms and more than 1 million lemmas. The MorfFlex CZ dictionary serves as an essential resource for ensuring the consistency of manual morphological annotation in the Prague Dependency Treebanks and underpins state-of-the-art automatic tools such as MorphoDiTa. 
In this paper, we focus on: (i) presenting an effective method for managing the rich morphological system within the dictionary, and (ii) demonstrating the utility of such a language resource for maintaining annotation consistency in corpora and supporting the development of advanced NLP applications. 
 \\ \newline  }

\usepackage{fancyhdr}
\fancypagestyle{officialbibref}{\fancyhf{}\fancyheadoffset[L,R]{50pt}\fancyhead[C]{This paper was published at \textbf{LREC 2026} -- please cite the published version {\small\url{https://aclanthology.org/2026.lrec-1.899/}}.}\fancyfoot[C]{\normalsize\thepage}\addtolength{\topmargin}{-50pt}\addtolength{\headsep}{50pt}}
\begin{document}
\thispagestyle{officialbibref}
\pagenumbering{arabic}\pagestyle{plain}

\maketitleabstract
\section{Introduction}

Over the course of many years of annotation work, numerous valuable linguistic resources (corpora and dictionaries) have been developed, forming the foundation for a wide range of successful applications and tools in the field of NLP. However, today the future of linguistic annotation lies at the intersection of traditional computational linguistics and modern approaches to natural language processing and artificial intelligence. The role of manual linguistic annotation in application development is steadily diminishing, and annotated resources are required (if at all) only in exceptional cases. This naturally raises the question of whether manual annotation still has a real purpose in the era of large language models. Drawing on several decades of experience with the development of linguistically annotated resources for both theoretical and applied research, we remain firmly convinced that a formalized, reusable source of linguistic knowledge is still a highly desirable goal of computational linguistics. 

\begin{table}[t!]
\small
\begin{center}
\begin{tabular}{l|l|l}
Wordform & Lemma & Tag \\\hline
{\it létat} \tiny{‘to-fly’}  & {\tt létat} & \verb|Vf--------A-I--|\\
{\it létám} \tiny{‘I-am-flying’}  & {\tt létat} & \verb|VB-S---1P-AAI--|\\
{\it nelétám} \tiny{‘I-not-am-flying’}  & {\tt létat} & \verb|VB-S---1P-NAI--|\\
{\it létáš} \tiny{‘you-are-flying’}  & {\tt létat} & \verb|VB-S---2P-AAI--| \\
{\it létá} \tiny{‘he-is-flying’}  & {\tt létat} & \verb|VB-S---3P-AAI--| \\
{\it létáme} \tiny{‘we-are-flying’}  & {\tt létat} & \verb|VB-P---1P-AAI--| \\
{\it létáte} \tiny{‘you-are-flying’}  & {\tt létat} & \verb|VB-P---2P-AAI--| \\
{\it nelétají} \tiny{‘they-are´nt-flying’}  & {\tt létat} & \verb|VB-P---3P-NAI--| \\
{\it nelétaj} \tiny{‘they-are´nt-flying’}  & {\tt létat} & \verb|VB-P---3P-NAI-6| \\
{\it létal} \tiny{‘he-flew’}  & {\tt létat} & \verb|VpYS----R-AAI--| \\
{\it létali} \tiny{‘they.Masc-flew’}  & {\tt létat} & \verb|VpMP----R-AAI--| \\
{\it létaly} \tiny{‘they.Fem-flew’}  & {\tt létat} & \verb|VpTP----R-AAI--| \\
{\it létaly} \tiny{‘they.Neut-flew’}  & {\tt létat} & \verb|VpNP----R-AAI-6| \\
{\it létala} \tiny{‘she-flew’} & {\tt létat} & \verb|VpQW----R-AAI--| \\
{\it létej} \tiny{‘fly!’} & {\tt létat} & \verb|Vi-S---2--A-I--|  \\
{\it létejme} \tiny{‘let's fly!’} & {\tt létat} & \verb|Vi-P---1--A-I--|  \\
{\it létáno} \tiny{‘flown’} & {\tt létat} & \verb|VsNS----X-API--|  \\
\end{tabular}
\caption{\textbf{MorfFlex CZ}: a sample of {<wordform, lemma, tag>} triplets for the paradigm \textit{létat} ‘to-fly’. The complete paradigm contains 106 triplets and 94 unique wordforms (some wordforms are homonymous, e.g. the wordform \textit{létaly} in this sample).} 
\vspace*{-0.15in}
\label{tab:example-of-triplets}
\end{center}
\end{table} 

\looseness-1
In this contribution, we present \textbf{MorfFlex CZ},\footnote{\url{https://ufal.mff.cuni.cz/morfflex}}
a morphological dictionary of Czech, which is a highly inflectional language. The dictionary was originally developed by Jan Hajič in the late 1980s \cite{hajic2004} as a basis for a spelling checker. Later, it was adapted for commercial applications, particularly lemmatization to support more efficient text search \citep{hajic-drozd-1990}. 
The dictionary then served as a key resource for manual morphological annotation of 1.7 million tokens in the Prague Dependency Treebank, a pioneering initiative and one of the very first treebanks and tools ever created. Even before the official release %
in 2001 \cite{pdt2001}, an earlier version, annotated at both the morphological and syntactic layers and comprising approximately 400,000 words, was used in 1998 at the Johns Hopkins University workshop. The corpus data provided training material for the first statistical taggers \citep{Hajic1998CoreNLP} and syntactic parsers \citep{collins-etal-1999-statistical,charniak-2000-maximum}. 

To date, the dictionary continues to be developed and manually maintained mainly as an essential resource to ensure the consistency of manual morphological annotation in the Prague Dependency Treebanks. The latest version is \textbf{MorfFlex CZ 2.1}\footnote{\url{http://hdl.handle.net/11234/1-5833}} \citeplanguageresource{morfflex}, 
fully compatible with the \textbf{Prague Dependency Treebank - Consolidated 2.0} annotation (\citealplanguageresource{pdtc20}; \citealp{pdtc-2026}). It also serves as the basis for state-of-the-art automatic tools such as MorphoDiTa (\citealp{strakova14}; see Sect.~\ref{use}).

\begin{table}[t!]
\begin{center}
\begin{tabular}{l|r}
\multicolumn{2}{c}{STATISTICS ON MORFFLEX CZ}\\\hline \hline
No. of unique lemmas & 1,058,079\\
No. of <wordform, lemma, tag> &  126,906,921 \\
\end{tabular}
\caption{Number of unique lemmas and <wordform, lemma, tag> triplets in MorfFlex CZ 2.1.}
\label{tab:statistics}
 \end{center}
\end{table}

The architecture of MorfFlex CZ is specially designed for inflectional languages with a large number of endings (suffixes). We refer to this architecture --- later applied to other languages with rich inflection (see Sect.~\ref{use}) --- simply as MorfFlex.

In inflective languages, words take endings to mark linguistic cases, grammatical number, gender, tense, etc. Therefore, many wordforms may be related to one lemma. E.g., the Czech word \textit{létat} (‘to-fly’) can appear as \textit{létám} ‘I-am-flying’, \textit{nelétám} ‘I-am-not-flying’,\footnote{There is also prefix \textit{ne-} ‘not/ir-/un-’, which regularly forms the negative wordforms.} \textit{létáš} ‘you-are
-flying’, \textit{letáme} ‘we-are-flying’, \textit{létejme} ‘let's-fly’, \textit{létalo} ‘it flew’, \textit{létali} ‘they flew’, etc. – there are several tens of wordforms for this type of verb. Corpus-wise (in the Prague Dependency Treebank – Consolidated 2.0 release), there are 230K unique wordforms and 97K lemmas in a corpus of almost 4M words (cf. Table~\ref{tab:corpus-wise}). It is therefore crucial to handle endings effectively and to reduce processing costs wherever regularities are found. The dictionary is distributed as a plain unstructured list of the \textbf{<wordform, lemma, tag>} triplets (as shown in Table~\ref{tab:example-of-triplets}). However, the manually maintained, unpublished source files and conversion scripts in fact represent a sophisticated system of inflectional and derivational patterns and rules which reduce the otherwise enormous size of the dictionary. Today, the dictionary contains more than 100 million wordforms and more than 1 million lemmas (cf. Table~\ref{tab:statistics}). We release the tool for expanding the source format into the basic <wordform, lemma, tag> triples under an open-source license at {\small\url{https://github.com/ufal/morfflex-generator}}.

In the paper, we describe the Czech morphological dictionary MorfFlex CZ and we focus on presenting:

\begin{itemize}
    \item  an effective method for handling rich morphological system of inflectional languages in the dictionary,  
    \item  the usefulness of such a language resource for ensuring the consistency of
corpus morphological annotation and for developing NLP applications.
\end{itemize}

The paper is organized as follows. In Sect.~\ref{related}, the related context of computational morphology is briefly outlined. Sect.~\ref{morfflex} provides an introduction to the system of handling rich morphology in MorfFlex. In Sect.~\ref{sec.formats}, the formats of the system is described in more detail: the unpublished source format (Sect.~\ref{sec.source}), the intermediate format (Sect.~\ref{sec.inter}), and the basic format (Sect.~\ref{sec.basic}) in which the dictionary is distributed. Sect.~\ref{sec.rules} describes the procedure for converting from the source format to the intermediate format (Sect.~\ref{derivationalrules}) and then to the basic format (Sect.~\ref{flection}). The direct path %
is described in Sect.~\ref{trivial}. The use of the dictionary is described in Sect.~\ref{use}. We conclude in Sect.~\ref{conclusion}.

\begin{table}[h]
\begin{center}
\begin{tabular}{l|r}
\multicolumn{2}{c}{CORPUS-WISE STATISTICS}\\\hline \hline
No. of tokens in corpus & 3,885,591 \\
No. of unique wordforms & 230,448\\
No. of unique lemmas & 97,101\\
\end{tabular}
\caption{Number of unique wordforms and lemmas in the Prague Dependency Treebank - Consolidated 2.0 release.}
\label{tab:corpus-wise}
 \end{center}
\end{table}

\section{Related Work}
\label{related}

In the context of NLP, morphology is primarily employed in the development of tools that recognize relationships between wordforms (morphological analyzers, e.g. \textit{Voikko},\footnote{\url{https://voikko.puimula.org/}} \textit{Hunmorph} \cite{hun-magyar}) or between related words (derivative tools, e.g. \citealp{derivace}). Many of these tools were originally created for spellchecking purposes %
(cf.  English \textit{SCOWL} (Spell Checker Oriented Word Lists)).\footnote{\url{http://wordlist.aspell.net/dicts/}}
For languages with a rich inflectional structure, for which a comprehensive list would be impractical, a combination of lemmas/roots and inflections is used (cf. \textit{Collatinus}\footnote{\url{https://outils.biblissima.fr/en/collatinus-web/}} for Latin,  \textit{Ajka}\footnote{\url{https://nlp.fi.muni.cz/projects/ajka/}} for Czech, or \textit{Polimorf}\footnote{\url{https://zil.ipipan.waw.pl/PoliMorf}} for Polish; for detailed description see e.g. \citealp{rectina}, \citealp{Osolsobe1996}, \citealp{latvian}). Morphological dictionaries are often created simultaneously with the compilation of a corpus, using manual or semi-manual annotation, for Slavic languages see e.g. \citealp{Sloleks-slovinci} and \citealp{chorvat}.
Efforts are also being made to unify morphological features and create a consistent approach to morphology (e.g.
\textit{Unimorph}\footnote{\url{https://unimorph.github.io}} \cite{batsuren-etal-2022-unimorph}; \textit{Paralex}\footnote{\url{https://www.paralex-standard.org/}} \cite{BeniamineEtAl2023Paralex}).

\section{Morphological Dictionary MorfFlex}
\label{morfflex}

The MorfFlex dictionary is a list of <wordform, lemma, tag> triplets. The lemma\footnote{Though lemma is often considered an abstract object, in MorfFlex it is always expressed as a human readable word.} is the basic form of the wordform (usually such a form that is used as an entry in general dictionaries). The tag codes morphological properties of the wordform.
The set of all wordforms with the same lemma is called a paradigm.
The lemma is usually viewed as a representative of the whole paradigm.
For example, the English wordform {\it fly} belongs to the paradigm represented by the lemma {\it fly}. 
The whole paradigm associated with the lemma {\it fly} is the set \{{\it fly, flies, flew, flying, flown}\}. 
Czech paradigms are usually much larger because of the rich inflection of Czech.
The paradigm of the Czech equivalent of the English {\it fly}, namely the verb {\it létat} has 94 unique wordforms in the MorfFlex CZ dictionary (including 30 rare archaic wordforms; cf. Table~\ref{tab:example-of-triplets}).

The set of triplets must comply with the so-called \textit{Golden rule of morphology} \cite{biblio:HlGoldenRule2017,novymanual}, which says that any particular pair <lemma, tag> can appear only in one triplet throughout the entire dictionary.
In other words, it is not possible that a pair <lemma, tag> can be used for the description of more than one wordform. It guarantees that generating wordforms from <lemma, tag> pairs is unambiguous. 
We apply tag numbering to distinguish between different types of wordform variants (cf. standard wordform \textit{nelétají} ‘they-are-not-flying’ and non-standard wordform \textit{nelétaj} ‘they-are-not-flying’ in Table~\ref{tab:example-of-triplets}).

The triplets of the morphological dictionary can be used for generating as well as analyzing wordforms.
On the basis of the pair <lemma, tag>, a single wordform is generated, as already explained. The opposite task, analysis, assigns a set of pairs <lemma, tag> to a given wordform.
In the latter case, there can be (often are) more such pairs, as the Czech language is massively homonymous. E.g., the morphological analysis assigns two pairs <lemma, tag> to the wordform {\it létaly} (cf. Table~\ref{tab:example-of-triplets}).

What we want to present here is that for Czech and other highly inflectional languages with a high degree of regularity (in inflection and/or in derivation), it is possible to describe paradigms using patterns -- sets of rules for the automatic generation of entire paradigms (in the form of triplets). %
The usage of patterns substantially reduces the size of the dictionary. It is also much more comprehensible for human maintainers. 

\begin{figure*}[ht!]
  \begin{center}
  \includegraphics[scale=0.6]{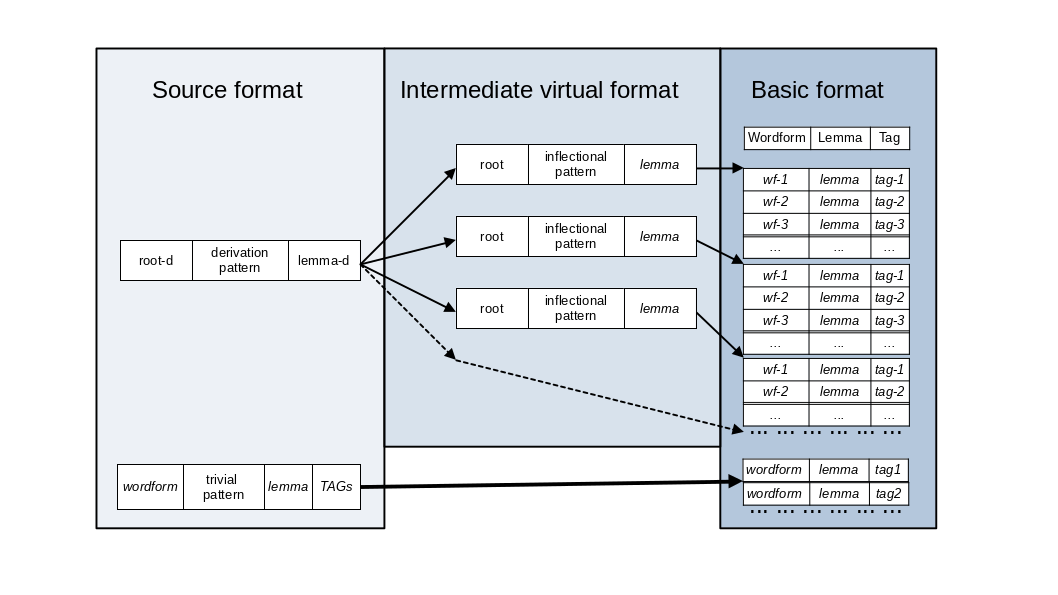}
 \caption{\textbf{MorfFlex scheme}. There are three formats: source, intermediate and basic. 
    In the \textbf{source format}, two types of records are used. The upper type contains a derivational pattern. According to specific rules for derivational patterns, this record is transformed into one or more records of the {\bf intermediate virtual format}, schematically drawn in the middle column. The derivational pattern is replaced by a set of inflectional patterns, the new roots are created and also the lemma is changed. Each of the records from the intermediate format generates a set of <wordform, lemma, tag> triplets -- they are shown in the rightmost part of the scheme, the final {\bf basic format}. 
    The lower part of the scheme shows the simplified procedure for trivial patterns. In that case, the record in the intermediate format would be the same as the source one, which means skipping the intermediate format.
    The italics font throughout the whole scheme represents the final items (strings), as they are being created while producing the basic format triplets.}
  \label{fig:all_formats}
  \end{center}
\end{figure*}

For various applications, the most convenient format of the morphological dictionary is the format of  <wordform, lemma, tag> triplets.  We call it the {\bf basic format}. On the other hand, the most convenient format for storing and maintaining the dictionary is the format with patterns, so-called {\bf source format}. Generating the basic format from the source format is not straightforward. In most cases, every record of the source format is first transformed into a set of records of a {\bf virtual intermediate format}, and subsequently into the basic format. This transformation is carried out using rules that are triggered by patterns in both the source format and the intermediate format. The whole procedure is presented as a schema in Fig.~\ref{fig:all_formats}. The patterns in the source format are \textbf{derivational}.\footnote{An exception is formed by the so-called trivial patterns (see Sect.~\ref{trivial}).} They are used for: 

\begin{enumerate}
    \item creating derivations (new words), namely their lemmas
    \item assigning inflectional patterns to every derived lemma
\end{enumerate}

The patterns in the intermediate format are  \textbf{inflectional}. They are used for: 
\begin{enumerate}
    \item generating wordforms associated with lemma
    \item assigning the tag describing the morphological properties of every wordform
\end{enumerate}

It is thus not necessary to include some (actually, many) words in the dictionary; they are derived using regular patterns for large sets of words. For instance, there is one  record in the source format for the verb {\it létat} ‘to fly’ (cf. the first row in Table~\ref{tab:source-format}). Its derivational pattern ({\tt ATN}) is ``translated'' into 20 records of the intermediate format (cf. sample in Table~\ref{tab:intermediate-format}). Here, an inflectional pattern is assigned to each of new derived lemmas. This procedure enables not only to create the whole paradigm of the originating verb {\it létat} ‘to fly’ itself, but also paradigms of the derived words such as {\it létávat} (‘to use to fly’), {\it létání} (‘flying’ -- noun), {\it létající} (‘flying’ -- adjective), etc. Thus, from the single line in the source format of the verb {\it létat} ‘to fly’, 3,096 triplets of the basic format are automatically  generated (cf. sample in Table~\ref{tab:example-of-triplets}).

\section{MorfFlex Formats}
\label{sec.formats}

In this section, all three MorfFlex formats are described: source (Sect.~\ref{sec.source}), intermediate (Sect.~\ref{sec.inter}) and basic (Sect.~\ref{sec.basic}).

\subsection{Source format}
\label{sec.source}

The source format of MorfFlex is stored in a text format in UTF-8 coding.
It enables very easy maintenance in any text editor. Every source format record of the dictionary contains the following pieces of information (prefix d- is for ''derivational''):

\begin{itemize}
    \item {\bf d-root}: beginning of a wordform that does not change within the d-pattern. 
    It does not need to be a grammatical root of a word. The d-root always includes prefixes that appear at the beginning of the wordform, except for the regular negation prefix {\it ne-}
    and the superlative prefix {\it nej-}.

    \item {\bf d-pattern}: derivational pattern (new lemmas are derived according to the rules stored in this pattern). %
    If it equals to {\tt 0} or {\tt 0n}, we call it \textit{trivial} pattern and in that case, there must be at least one item TAGs present in the dictionary source line.
    In this paper, the d-patterns are written in capital letters, contrary to the inflectional patterns.
    
    \item {\bf d-lemma}: originating lemma. This item can be enriched with several types of information concerning variants, style, or just an explanation for maintainers. 
    They do not have any influence on the procedure of generating the wordforms and their tags.
    In case of homonymy, the lemmas are numbered to represent different paradigms, e.g., {\tt jak-1} is for noun (‘yak’~-~animal), {\tt jak-2} is for conjunction (‘as’), {\tt jak-3} is for adverb (‘how’). See Table~\ref{tab:source-format}. 
    
    \item {\bf TAGs}: morphological tags assigned to the wordform. It is present only when d-pattern is trivial (e.i., equals {\tt 0} or {\tt 0n}). %
\end{itemize}

Many paradigms are described by more records in the source format. In that case, the d-lemma is identical in all records belonging to that paradigm. Typical examples are verbs with irregular conjugation. Also, irregular or atypical inflection of nouns is described using several records, usually by adding a record with the trivial pattern %
(cf. record of {\it ocet} ‘vinegar’ in Table~\ref{tab:source-format}).
Examples of the source format records are shown in Table~\ref{tab:source-format}.

\begin{table} [t!]
\begin{center}
\small
\begin{tabular}{l|l|l|l}
 {d-root} & {d-pattern} & {d-lemma} & {TAGs} \\
\hline 
lét & ATN & létat & \\ 
jak & PN18 & jak-1\\
jak & 0 & jak-2 & {\verb|J,-------------|} \\
jak & 0 & jak-3 & {\small \verb|Db-------------|} \\
oc & HD1ET & ocet &   \\
octa & 0 & ocet & {\small \verb|NNIS2-----A---1|}  \\
octě & 0 & ocet & {\small \verb|NNIS6-----A---1|}  \\
\end{tabular}
\caption{\textbf{Record examples in the source format}. The first row is examples of record verb {\it létat} ‘to-fly’ with derivational pattern. The following 3 rows show records of the homonymous lemma {\it jak}  - one with regular d-pattern; the rest shows the lemma {\tt ocet} 'vinegar' with regular d-pattern and two more trivial patterns {\tt 0} for the description of two irregular wordforms, not covered by the d-pattern.}
\label{tab:source-format}
\end{center}
\end{table}

\subsection{Intermediate format}
\label{sec.inter}

The intermediate format is only virtual. It represents a transitional stage in the overall process of generating the final triplets, from the source format to the basic one. The intermediate format records has the same items (except TAGs) as the records in the source format, but their content is different: instead of derivational patterns, there are inflectional patterns. Thus, a virtual intermediate record contains the following pieces of information (prefix i- is for ''inflectional''):

\begin{itemize}
    \item \textbf{i-root}: beginning of a wordform that does not change during the inflection within the i-pattern (used for the creation of the whole paradigm for i-lemma)
    \item \textbf{i-pattern}: inflectional pattern. %
    \item \textbf{i-lemma}: inflectional lemma -- the final lemma that appears in the basic format. The newly derived i-lemmas include a so-called \textbf{\_backlink} to the originating lemma. 
\end{itemize}

Examples of intermediate format records are shown in Table~\ref{tab:intermediate-format}.

\begin{table}[t]
 \begin{center}
 \small
\begin{tabular}{l|l|l|l}
i-root & i-pattern & i-lemma & \_backlink \\
\hline
lét & atn & létat \tiny{‘to fly’} & \\ 
létajíc & iavg & létající \tiny{‘flying - ADJ’} & \_létat  \\
létání & stn & létání \tiny{‘flying - NOUN’} & \_létat \\
létáv & atn & létávat \tiny{‘to-use-to fly’} &  \_létat \\
létávání & stn & létávání \tiny{‘use-to-fly-NOUN’} &  \_létat \\
\end{tabular}
\caption{Examples of several \textbf{records in the intermediate format} derived from the source format record of the d-lemma \textit{létat} ‘to fly’ (cf. the first row in Table~\ref{tab:source-format}).} 
\label{tab:intermediate-format}
\end{center}
\end{table}

\subsection{Basic format}
\label{sec.basic}

The basic format of MorfFlex is a plain unstructured list of  <wordform, lemma, tag> triplets:

\begin{itemize}
    \item \textbf{wordform}
    \item \textbf{lemma} - representative wordform. Lemma is unique for the whole paradigm, i.e. a set of wordforms grouped according to inflectional behaviour.
    \item  \textbf{tag} - In the positional tag, full inflectional information is coded for the wordform. An overview and explanation of the tag positions are provided in Table~\ref{tab:tag}.
    \begin{table}[h]
\begin{center}
\small
\begin{tabular}{r|l}
Position & Description \\\hline
1 & Part of speech \\
2 & Detailed part of speech \\
3 & Gender \\
4 & Number \\
5 & Case \\
6 & Possessor's gender \\
7 & Possessor's number \\
8 & Person \\
9 & Tense \\
10 & Degree of comparison \\
11 & Negation \\
12 & Voice \\
13 & Verbal aspect \\
14 & Aggregate \\
15 & Variant, style, abbreviation
\end{tabular}
\caption{Attributes in positional tags}
\label{tab:tag}
\end{center}
\end{table} 
\end{itemize}

An example of the final basic format is shown in Table~\ref{tab:example-of-triplets}. It displays the final triplets generated to the basic format based on the intermediate record shown in the first row of Table~\ref{tab:intermediate-format}.

\section{MorfFlex Procedure}
\label{sec.rules}

The procedure for creating a dictionary from the source format is not straightforward; it is carried out in two steps. 
First, the source format is converted into a virtual intermediate format (Sect.~\ref{derivationalrules}), from which the basic format is subsequently generated (Sect.~\ref{flection}). 
Those source format records, that contain a trivial pattern (and therefore skip over the intermediate format) are used for generation the triplets of the basic format directly (Sect.~\ref{trivial}).

\subsection{From source to intermediate format}
\label{derivationalrules}

In this section, we describe the procedure of the transformation from the source format (described in Sect.~\ref{sec.source}) into the intermediate format (Sect.~\ref{sec.inter}). 

Every source format record (except those with trivial patterns) is transformed into one or more intermediate records according to the set of special derivational rules. This means that the d-root in the source format can be different from the i-root(s) of the intermediate format, and the d-lemma is changed into a possibly different i-lemma(s). For each new i-root and i-lemma, an inflectional i-pattern is set.

There are one or more derivational rules for each d-pattern.  Every derivational rule is represented by a single line that codes the actions to be done with a record of the source format to create the record(s) in the intermediate format. This is a slightly simplified format of a derivational rule:

\begin{center}
    {\tt d-pattern r1,r2,r3,r4,r5,r6}  %
\end{center}

The explanation is as follows:
\begin{itemize}
    \item {\bf d-pattern} is a derivational pattern from the source format of the dictionary.
    \item {\bf r1} is a string that is to be concatenated with the d-root. The result is a new root (i-root). 0~(zero) means an empty string.
    \item {\bf r2} is an inflectional pattern that is to be used to generate the paradigm of the new \mbox{lemma}.
    \item {\bf r3} codes the first step of the construction of the new derived lemma (i-lemma).%
    \item {\bf r4} is the ending of the new derived lemma. %
   \item {\bf r5} and {\bf r6} relate to a backlink to the originating d-lemma from which the new i-lemma was created --- a derivational link.  The derivational link then becomes part of the new lemma, in a form that is rather a note to hint which derivation took place.
\end{itemize}

The entire procedure of using derivational rules is demonstrated in detail on an example in Sect.~\ref{example-sc1}.

\subsubsection{Example of using a derivational rule}
\label{example-sc1}

For the demonstration of the derivational rules, we chose the example {\it dárce} ‘donor’ with the derivational pattern {\tt SC1}. The source format record is shown in Table~\ref{tab:sc1}.

\begin{table}[t]
 \begin{center}
 \small
\begin{tabular}{l|l|l|l}
d-root & d-pattern & d-lemma & TAGs\\
\hline
dár & SC1 & dárce &  \\
\end{tabular}
\caption{Source format record for d-lemma \textit{dárce} ‘donor’.}
\label{tab:sc1}
\end{center}
\end{table}

The derivational rules for the (derivational) d-pattern {\tt SC1} are in Table~\ref{tab:sc1rules}. Each of the four lines in Table~\ref{tab:sc1rules} is used to derive one record of the intermediate format as presented in Table~\ref{tab:donor2}. It summarizes all the derived lemmas and their inflectional patterns that will generate the paradigm triplets in the basic format. The rows in Table~\ref{tab:donor2} are numbered following the order of the derivational rules in Table~\ref{tab:sc1rules}. 

\begin{table}[t]
\centering
\small
\begin{tabular}{l|l|l}
 & d-pattern  & rule \\ \hline
        1. & {\tt SC1}  & 0,sc1,0,0,0,0 \\
        2. & {\tt SC1} & c,uv,r0,cův,0,0\\
        3. & {\tt SC1} & ky,ns3n,r0,kyně,0,0\\
        4. & {\tt SC1} & kynin,in,r0,kynin,r0,kyně\\
    \end{tabular}
    \caption{Derivational rules for the {\tt SC1} d-pattern.}
    \label{tab:sc1rules}
\end{table}

\begin{table}[t]
\begin{center}
\small
\begin{tabular}{l|l|l|l}
& i-root & i-pattern & i-lemma\_back-link\\
\hline
1 & dár & sc1 & dárce \\
2 & dárc & uv & dárcův\_dárce \\
3 & dárky & ns3n & dárkyně\_dárce \\
4 & dárkynin & in & dárkynin\_dárkyně \\
\end{tabular}
\caption{Records of the intermediate format derived from the source format record presented in Table~\ref{tab:sc1} --
according to derivational rules presented in Table~\ref{tab:sc1rules}.
The meaning of the i-lemmas are: 1~donor, 2 belonging to donor, 3 donor-woman, 4 belonging to donor-woman.}
\label{tab:donor2}
\end{center}
\end{table}

Based on a single line record in the source format (Table~\ref{tab:sc1}), four different virtual intermediate records (Table~\ref{tab:donor2}) are created according to derivational rules presented in Table~\ref{tab:sc1rules}. Assigned inflectional i-patterns are to be used in the next step (cf. Sect.~\ref{example-i}) for the generation of 250 different triplets in the basic format of the dictionary.%

\subsection{From intermediate to basic format}
\label{flection}

The intermediate format of the dictionary (Sect.~\ref{sec.inter}) is transformed into the basic format (Sect.~\ref{sec.basic}) by means of inflectional i-patterns that appear in records of the intermediate format. The inflectional i-patterns contain a prescription for generating the wordforms and tags of the basic format triplets, while the lemma is simply copied from the intermediate virtual record.

Every i-pattern consists of a set of pairs <ending, tag>, associating the ending (it is simply a string to be concatenated with the i-root) with a particular set of grammatical morphological categories represented by the tag. 

The entire procedure of using an inflectional rule is demonstrated on an example in the next Sect.~\ref{example-i}.

\subsubsection{Example of using an inflectional rule}
\label{example-i}

To demonstrate the usage of inflectional rules, we chose the intermediate record for the lemma \textit{dárce} ‘donor’ with the i-pattern {\tt sc1}. The intermediate format record is shown in Table~\ref{tab:sc1-i} (cf. also the first line of Table~\ref{tab:donor2}).

\begin{table*}[t]
    \centering
\small
\begin{tabular}{|l||l|l||l|l|}
\hline
\multicolumn{4}{|c|}{sc1} \\ \hline
 end & tag &  end & tag \\ \hline
-ci & \verb|NNMP1-----A----| &  -cema & \verb|NNMP7-----A---6|\\ \hline
 -cové & \verb|NNMP1-----A---1|  & -ce & \verb|NNMS1-----A----|\\ \hline
 -ců & \verb|NNMP2-----A----| &  -ce & \verb|NNMS2-----A----|\\ \hline
 -cům & \verb|NNMP3-----A----| &  -ci & \verb|NNMS3-----A----|\\ \hline
 -cum & \verb|NNMP3-----A---6| &  -covi & \verb|NNMS3-----A---1|\\ \hline
 -ce & \verb|NNMP4-----A----| &   -ce & \verb|NNMS4-----A----|\\ \hline
 -ci & \verb|NNMP5-----A----| &   -ce & \verb|NNMS5-----a----|\\ \hline
 -cové & \verb|NNMP5-----A---1| & -ci & \verb|NNMS6-----A----|\\ \hline
 -cích & \verb|NNMP6-----A----| & -covi & \verb|NNMS6-----A---1|\\ \hline
 -ci & \verb|NNMP7-----A----| &   -cem & \verb|NNMS7---------|\\ \hline
\end{tabular}
    \caption{Example of the inflectional {\tt sc1} i-pattern,  according to which, for example, the paradigm  of the noun \textit{dárce} ‘donor’ is generated.}%
    \label{tab:i-pattern-sc1}
\end{table*}

\begin{table}[t!]
 \begin{center}
 \small
\begin{tabular}{l|l|l}
i-root & i-pattern & i-lemma\\
\hline
dár & sc1 & dárce \\
\end{tabular}
\caption{Intermediate record for i-lemma \textit{dárce} ‘donor’.}
\label{tab:sc1-i}
\end{center}
\end{table}

The inflectional rules associated with i-pattern {\tt sc1} are demonstrated in Table~\ref{tab:i-pattern-sc1}. These rules combine the i-root \textit{dár} with the endings (end) and tags in the Table~\ref{tab:i-pattern-sc1}, and generate the resulting basic format triplets (a sample is shown in Table~\ref{tab:example-of-darce}).

\begin{table}[t!]
\small
\begin{center}
\begin{tabular}{l|l|l}
Wordform & Lemma & Tag \\\hline
{\it dárci} \tiny{‘donors-Nominative’}  & {\tt dárce} & \verb|NNMP1-----A----|\\
{\it dárcové} \tiny{‘donors-Nomin.’}  & {\tt dárce} & \verb|NNMP1-----A---1|\\
{\it dárců} \tiny{‘donors-Genitive’}  & {\tt dárce} & \verb|NNMP2-----A----|\\
{\it dárcům} \tiny{‘donors-Dative’}  & {\tt dárce} & \verb|NNMP3-----A----| \\
{\it dárce} \tiny{‘donors-Accusative’}  & {\tt dárce} & \verb|NNMP4-----A----| \\
{\it dárcové} \tiny{‘donors-Vocative’}  & {\tt dárce} & \verb|NNMP5-----A----| \\
{\it dárcích} \tiny{‘donors-Locative’}  & {\tt dárce} & \verb|NNMP6-----A----| \\
{\it dárci} \tiny{‘donors-Instrumental’}  & {\tt dárce} & \verb|NNMP7-----A----| \\
\end{tabular}
\caption{A shortened sample of {<wordform, lemma, tag>} triplets for the paradigm \textit{dárce} ‘donor’.} 
\label{tab:example-of-darce}
\end{center}
\end{table}

\subsection{Trivial patterns}
\label{trivial}

In the case of trivial patterns, we can interpret the whole procedure as skipping over the intermediate format and generating the basic format triplets directly from the source format (see  Fig.~\ref{fig:all_formats}). There are only two trivial patterns, {\tt 0} and {\tt 0n}, the latter differs from the former by its ability to generate also a negative wordform, which is formed by placing the prefix {\it ne-} ‘non/ir/un’ at the beginning of the wordform. %
No other changes to the root are induced by trivial patterns. Every source record with a trivial pattern must contain the item TAGs, that consists of a set of morphological tags. For each tag, one triplet is generated directly in the basic format. The wordform equals the d-root and the lemma equals the d-lemma.  

The trivial patterns are mostly used in case of fully or partially irregular paradigms, but also for uninflected forms. %
For fully irregular paradigms, the whole paradigm has to be described as a set of records with the trivial patterns, form by form. 
Partially irregular paradigms are described by means of non-trivial pattern(s) plus additional record(s) with a trivial pattern (cf. e.g. source format records for the d-lemma \textit{ocet} ‘vinegar’ in Table~\ref{tab:source-format}).

The procedure with trivial patterns is demonstrated on an example in the next Sect.~\ref{example-trivial}.

\subsubsection{Examples of using a trivial pattern}
\label{example-trivial}

To demonstrate the direct procedure with trivial patterns from source format to basic format, we chose the words {\textit{dnes} ‘today’} and {\textit{zřídka} ‘rarely’}. The former one has the trivial pattern {\tt 0}, the latter one {\tt 0n} enabling to create both affirmative and negative wordforms. See source format records in  Table~\ref{tab:example-tr}. The resulting triplets of the these records are in Table~\ref{tab:triplets-tr}. 

\begin{table}[h!]
 \begin{center}
\small
\begin{tabular}{l|l|l|l}
d-root & d-pat. & d-lemma & TAGs \\
\hline
dnes & 0 & dnes & {\small \verb|Db-------------|}  \\
zřídka & 0n & zřídka & {\small \verb|Dg-------1@----|}  \\
\end{tabular}
\caption{Source format records of adverbs {\textit{dnes} ‘today’} and {\textit{zřídka} ‘rarely’} with trivial patterns. The placeholder @ in the second record tag will be replaced with {\tt A} for the affirmative and {\tt N} for the negative wordform, see the following Table~\ref{tab:triplets-tr}.}
\label{tab:example-tr}
\end{center}
\end{table}

\begin{table}[h]
\small
\begin{center}
\begin{tabular}{l|l|l}
Wordform & Lemma & Tag \\\hline
{\it dnes}  & {\tt dnes} & \verb|Db-------------|\\
{\it zřídka}  & {\tt zřídka} & \verb|Dg-------1A----|\\
{\it nezřídka}  & {\tt zřídka} & \verb|Dg-------1N----|\\
\end{tabular}
\caption{The resulting {<wordform, lemma, tag>} basic format triplets for the paradigm  {\textit{dnes} ‘today’} and {\textit{zřídka} ‘rarely’} of the trivial source format records presented in Table~\ref{tab:example-tr}.
} 
\label{tab:triplets-tr}
\end{center}
\end{table} 

\section{Use of MorfFlex}
\label{use}

Morphological analysis, part-of-speech tagging are important components of NLP applications. They usually represent initial steps of language processing. Despite recent advances in part-of-speech and morphological tagging, the old truth that more data always gives better results \cite{banko-brill-2001-scaling,church-mercer-1993-introduction} still holds. At the same time, consistency in data annotation is a very important factor. For morphological annotation, especially for morphologically rich languages with thousands of possible combinations of morphological values, consistency can only be achieved when a dictionary lists all plausible morphological interpretations of all wordforms (cf. \citealp{hajic-2000-morphological}). 

The morphological dictionary MorfFlex CZ has from the very beginning been an essential part of the manual morphological annotation in the Prague Dependency Treebanks (the latest release is \textbf{Prague Dependency Treebank – Consolidated 2.0} (\citealplanguageresource{pdtc20}; \citealp{pdtc-2026}), containing almost 4M tokens manually annotated for morphology).\footnote{For details see a specification of the Czech morphological annotation \citep{novymanual}.} As the volume of manually annotated data increases, the dictionary is naturally being expanded and enriched as well. The aim of all annotation projects is to achieve full consistency not only between the data and the dictionary, but also also within both the data and the dictionary themselves (cf. \citealp{pdtc10-2020,modification-slovko2019}).

The PDT-C corpus and MorfFlex CZ dictionary are used for building models for \textbf{MorphoDiTa} (Morphological Dictionary and Tagger),\footnote{\url{https://lindat.mff.cuni.cz/services/morphodita}} an open-source tool for morphological analysis, generation, tokenization, lemmatization and tagging of texts. It performs morphological analysis and generation using the MorfFlex CZ dictionary. The MorphoDiTa tool achieves state-of-the-art results with a throughput around 10-200K words per second. It has, on average across datasets in PDT-C, F1-scores of 96.27\% for tagging and 98.31\% for lemmatization \cite{udpipe2024}.
MorphoDiTa is used for automatic POS and morphological annotation of all the Czech corpora available in the Kontext KWIC tool at the LINDAT/CLARIAH-CZ research infrastructure.\footnote{\url{https://lindat.cz}} 
MorphoDiTa is also one of the two components used for morphological disambiguation of the Czech National Corpus\footnote{\url{https://www.korpus.cz}} \cite{PetkevicJelinek2025}.

On the manual annotation of PDT-C data, the \textbf{Czech UDPipe model}, is trained. UDPipe\footnote{\url{https://lindat.mff.cuni.cz/services/udpipe}} is a pipeline for tokenization, tagging, lemmatization and dependency parsing. UDPipe took part in several competitions, reaching excellent results in all of them \cite{zeman-etal-2018-conll,mccarthy-etal-2019-sigmorphon,sprugnoli-etal-2020-overview}. In \cite{udpipe2024}, the authors show that a model, which combines the deep learning architecture of UDPipe with rescoring by the morphological dictionary MorfFlex (the core of MorphoDiTa), improves over both a deep learning system and a dictionary-based system on their own.

MorfFlex CZ is also aligned with the set of 1,040,126 lexemes contained in \textbf{DeriNet} \cite{DeriNet202019}, a lexical network which models word-formation relations in the lexicon of Czech. 

The MorfFlex framework can also be used for other languages with rich morphology. In addition to Czech, there is also a version for Slovak called \textbf{MorfFlex SK} \cite{morfflexSK}. This dictionary is also used in MorphoDiTa for the Slovak language.

\section{Conclusion}
\label{conclusion}
In conclusion, MorfFlex exemplifies how a systematically organized morphological dictionary can effectively manage the complexity of a highly inflective language such as Czech. By combining manually maintained source files with derivational and inflectional patterns, the dictionary achieves a remarkable balance between comprehensiveness and efficiency: from 450K rows in the source format, it produces over 100 million wordforms and more than 1 million lemmas. MorfFlex CZ ensures consistency in manual morphological annotation within the Prague Dependency Treebanks and also provides a robust foundation for state-of-the-art NLP tools like MorphoDiTa.

This contribution highlights the continued relevance of formalized linguistic resources in the era of advanced computational methods. MorfFlex demonstrates that even in highly inflectional languages, careful linguistic modelling can substantially reduce the size of a dictionary while ensuring it remains usable for human annotators and automated systems. Ultimately, this resource underscores the ongoing importance of combining linguistic expertise with computational techniques to advance corpus annotation, NLP development, and the broader study of morphologically rich languages.

\section{Acknowledgements}
The research and language resource work reported in the paper has been supported by the LINDAT/CLARIN and LINDAT/CLARIAH-CZ projects funded by Ministry of Education, Youth and Sports of the Czech Republic (projects LM2015071, LM2018101, LM2023062). The original annotation has been supported by multiple projects in the past, funded both nationally by the Ministry of Education, Youth and Sports of the Czech Republic and the Czech Science Foundation.

\section{Limitations}

The presented pattern-based management of rich morphological systems is limited to inflectional languages (primarily Slavic ones). For languages of an agglutinative type, it would need to be adapted. On the one hand, the pattern-based system makes it possible to efficiently generate a huge number of lemmas and wordforms; on the other hand, even a minor change in one pattern can lead to numerous changes in the resulting dictionary, which can be difficult to monitor.

\section{Bibliographical References}\label{sec:reference}

\bibliographystyle{lrec2026-natbib}
\bibliography{lrec2026-example}

\section{Language Resource References}
\label{lr:ref}
\bibliographystylelanguageresource{lrec2026-natbib}
\bibliographylanguageresource{languageresource}

\end{document}